\pdfoutput=1

\documentclass{article}
\usepackage{ijcai20}

\usepackage{times}

\usepackage{soul}
\usepackage{url}
\usepackage[hidelinks]{hyperref}
\usepackage[small]{caption}
\usepackage[pdftex]{graphicx}
\usepackage{amssymb}
\usepackage{amsmath}
\usepackage{gensymb}
\usepackage{bm}
\usepackage{booktabs}
\usepackage{subfigure}
\usepackage[ruled]{algorithm2e}
\usepackage[pdftex]{color}
\usepackage{multirow}

\newcommand{\zhihui}[1]{{\color{black}{#1}}}

\newcommand{\cloudModel}{target model\xspace}
\newcommand{\cloudModels}{target models\xspace}
\newcommand{\monitorModel}{monitor model\xspace}
\newcommand{\monitors}{monitor models\xspace}

\title{Increasing Trustworthiness  of Deep Neural Networks via Accuracy Monitoring\footnote{Accepted by the AISafety workshop co-located with IJCAI-PRICAI 2020. }}
\author{
Zhihui Shao\and
Jianyi Yang\and
Shaolei Ren
\affiliations
University of California, Riverside\\
\emails
zshao006@ucr.edu,
jyang239@ucr.edu,
sren@ece.ucr.edu
}
\begin{document}

\maketitle

\begin{abstract}
Inference accuracy of
deep neural networks (DNNs)
is a crucial performance metric, but can vary greatly in practice subject to actual test datasets and is typically unknown  due to the lack of ground truth labels. This has raised significant concerns with trustworthiness of DNNs, especially in safety-critical applications. In this paper, we address trustworthiness of DNNs by using  post-hoc processing to monitor
 the true inference accuracy on a user's dataset. Concretely,
we propose a neural network-based accuracy monitor model,
which only takes the deployed DNN's softmax probability output as its input 
and directly predicts if the DNN's prediction result is correct or not,
thus leading to an estimate of the true inference accuracy.
The accuracy monitor model can be pre-trained on
a dataset relevant to the target application of interest, and only needs to actively label a small portion (1\% in our experiments) of the user's dataset for model transfer. For estimation robustness, we further
employ an ensemble of monitor models based on the Monte-Carlo  dropout method.
We evaluate our approach on different deployed DNN models for image classification and traffic sign detection over multiple datasets (including adversarial samples). The result shows that our accuracy monitor model provides a
close-to-true accuracy estimation and outperforms the existing baseline methods.
\end{abstract}

\section{Introduction}

Deep neural networks (DNNs) have achieved unprecedentedly
high classification accuracy and found
success in numerous applications, including image classification, speech recognition,
and nature language processing.
Nonetheless, training an error-free or 100\% accurate DNN
is impossible in most practical cases. Inference
accuracy is a crucial metric for quantifying the performance of DNNs.
Typically, the reported inference accuracy of a DNN is measured offline
on test datasets with labels,
but this
 can significantly differ from the true accuracy
on a user's dataset because of, e.g., data distribution
shift away from the training dataset
or even adversarial modification to the user's data
\cite{Verification_DeepVerifier_OOD_Adversarial_Bengio_2019_arXiv,DNN_Uncertainty_PostHoc_Dirichlet_NIPS_2019_kull2019beyond,DNN_Uncertainty_PriorNetworks_NIPS_2018_malinin2018predictive}.
Moreover, obtaining the true accuracy is very challenging in practice
due to the lack of ground-truth labels.

The unknown inference accuracy has further decreased
the transparency of already hard-to-explain DNNs
and raised significant concerns with their trustworthiness, especially in safety-critical applications.
Consequently,
studies on increasing trustworthiness of DNNs
have been proliferating.
For example, many studies have
considered out-of-distribution (OOD) detection and
adversarial sample detection, since OOD and adversarial
samples often dramatically decrease inference accuracy of DNNs
\cite{DNN_Uncertainty_Baseline_OOD_ICLR_2017,Verification_DeepVerifier_OOD_Adversarial_Bengio_2019_arXiv,Verification_SimpleFramework_OOD_Generative_Gaussian_NIPS_2018_7947,Verification_OOD_Detection_ODIN_Srikant_UIUC_ICLR_2018}.
While these efforts can offer an increased assurance of DNNs
 to users to some extent,
they do not provide
a quantitative measure of actual classification accuracy, which
is a more direct and sensible measure
of the target DNN's performance.
 Some other studies propose (post-hoc) processing
to quantify/estimate
the prediction confidence of a DNN
\cite{DNN_Calibration_TemperatureScaling_ICML_2017_calibration_guo_2017,DNN_Uncertainty_PostHoc_Dirichlet_NIPS_2019_kull2019beyond,DNN_Uncertainty_Estimation_Summary_NIPS_2019_ovadia}.
Nonetheless, they typically require 
the target
DNN's  training/validation dataset
to train a (sometimes complicated)
new transformation model for confidence calibration,
and do not transfer well to new unseen datasets.
The accuracy of a target DNN on a user's operational dataset
can also be estimated via selective random sampling,
but it can suffer from a high estimation variance \cite{DNN_Operational_Testing_NJU_FSE_2019_10.1145/3338906.3338930}.

\begin{figure}[!t]
	\centering
	\includegraphics[width=0.48\textwidth, trim=0 12cm 11.5cm 0]{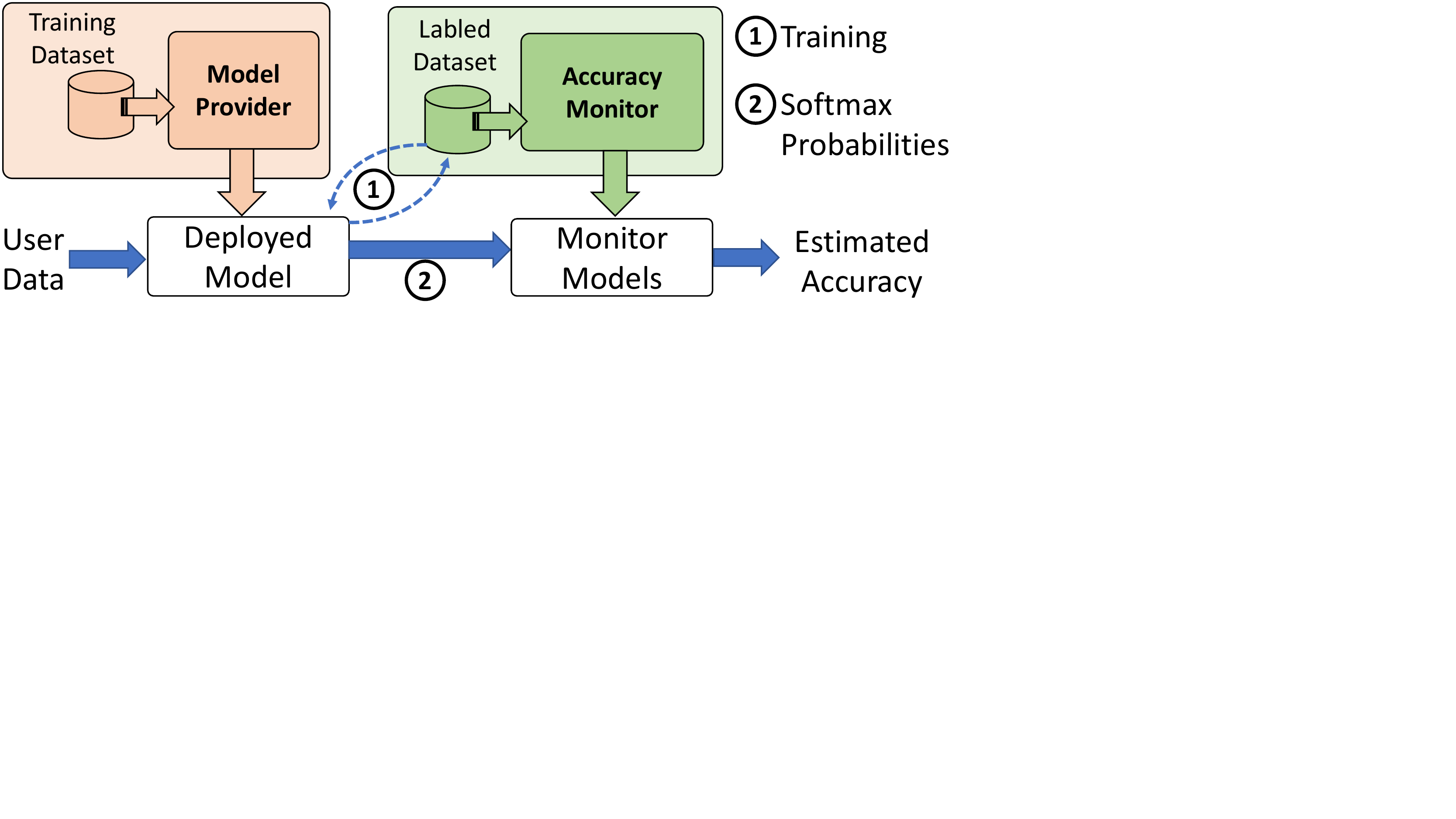}
	\vspace{-0.3cm}
	\caption{Accuracy monitoring for a deployed/target DNN.}\label{fig:illustration}
	\vspace{-0.3cm}
\end{figure}

\textbf{Contribution.} In this paper, we propose a simple yet effective post-hoc method ---
\emph{accuracy monitoring} --- which increases
the trustworthiness of DNN classification results
by estimating
the true inference accuracy on an actual (possibly OOD/adversarial) dataset.
Concretely, as shown in Fig.~\ref{fig:illustration}, we propose a neural network-based accuracy monitor model,
which only takes the deployed
DNN's softmax probability output as its input
and directly predicts if the DNN's prediction result is correct or not.
Thus, over a sequence of prediction samples from
a user's dataset, our accuracy monitor
can form an estimate of the target DNN's true inference accuracy.
Furthermore, we employ an
ensemble of monitoring models based on the Monte-Carlo  dropout method,
providing a robust
estimate of the target DNN's true accuracy.

Utilizing as little information as the target DNN's softmax probability output for accuracy estimation provides better transferability than
more complicated calibration methods \cite{DNN_Uncertainty_PostHoc_Dirichlet_NIPS_2019_kull2019beyond}.
Specifically, we can pre-train an accuracy monitor model based on
a labeled dataset relevant to the target application of interest
(e.g., public datasets for image classification).
Then, for model transfer, we can
selectively label a small amount (1\% in our work)
of data from the user's test dataset with active learning via
an  entropy acquisition function \cite{ensemble_active_learning_2018},
and re-train our monitor models
on the selectively labeled data using transfer learning.
In addition,
without the need
of accessing the target DNN's training/validation datasets,
our accuracy monitoring method
can be easily applied
as a plug-in module on top of the target DNN
to monitor its runtime performance on a variety
of datasets. Thus, our method is not restricted
to the DNN providers themselves; instead, even
an end user can employ our method to monitor
the target DNN's accuracy performance on its own, bringing
 further increased trustworthiness of accuracy
monitoring.

To evaluate the effectiveness of our accuracy monitoring method,
we consider different target DNN models for
image classification (10 classes and 1000 classes)
and for traffic sign detection in autonomous driving, respectively.
Our results show that, by only utilizing the prediction class and softmax probability output of the deployed DNN model and labeling 1\% of the user's dataset, our method can monitor the healthy of the target DNN models, providing a remarkably accurate estimation of the true classification accuracy on a variety of user's datasets.

\begin{figure*}[!t]
	\centering
	\includegraphics[width=0.95\textwidth, trim=0 9cm 3.8cm 0]{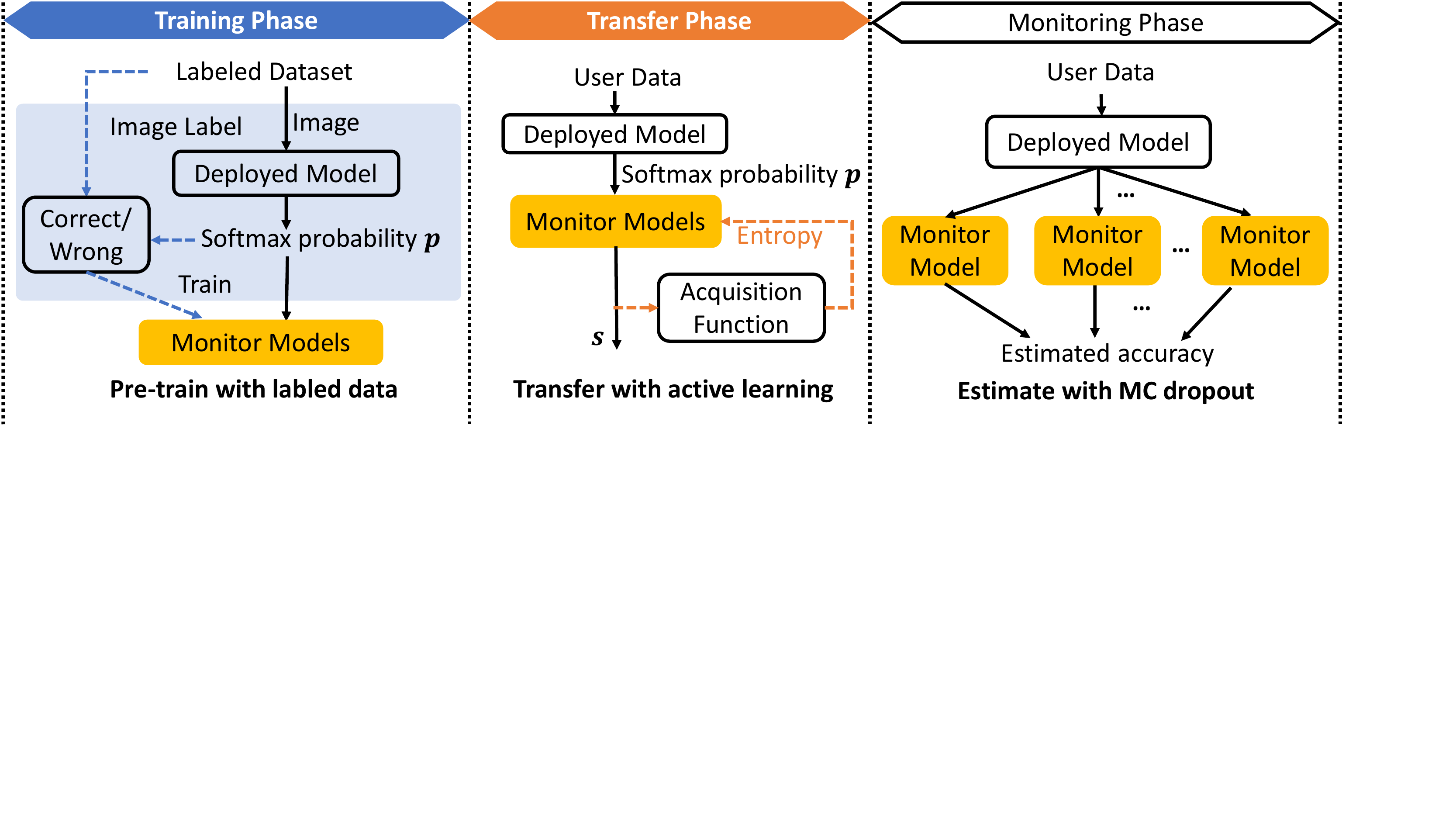}
	\vspace{-0.2cm}
	\caption{DNN accuracy monitoring: {Training}, transferring and accuracy estimation.}\label{fig:monitor_system_overview}
	\vspace{-0.2cm}
\end{figure*}

\section{Related Works}

\textbf{Prediction uncertainty estimation.}
Several methods have been proposed to estimate DNN prediction uncertainty.
 In \cite{model_uncertain_JohnsHopkins_2019}, the model uncertainty is estimated with ensemble models via re-sampling the original DNN model parameters based on the Hessian matrix and gradient matrix on the training data.
Additionally, \cite{classifier_uncertainty_google_nips_2018} estimates model uncertainty via the similarity between the test data and training data. However, it requires not only the training data but also a white-box target DNN model. Other methods (e.g, MC dropout, ensembles, stochastic variational Bayesian inference, prior networks) are summarized in \cite{DNN_Uncertainty_PriorNetworks_NIPS_2018_malinin2018predictive,DNN_Uncertainty_Estimation_Summary_NIPS_2019_ovadia}, which also require a white-box model and/or the original training dataset.
By contrast, our post-hoc
processing method only needs the target DNN's softmax
probability output and applies to a variety
of datasets, including OOD and adversarial samples.

\textbf{Concept/distribution drift detection.} After model deployment,
some studies indirectly tackle the problem
of model accuracy monitoring via concept/data distribution drift detection in the absence of labels.
In \cite{auto_model_monitor_pinto_2019}, an automatic concept drift detection algorithm \emph{SAMM} is developed with no labeled test data by utilizing the feature distance between test data and reference data. Other approaches include
 \emph{ML Health} \cite{ml_health_Ghanta_2019} and \emph{MD3}  \cite{detection_concept_drift_Sethi_2017}.
Moreover, \cite{Verification_DeepVerifier_OOD_Adversarial_Bengio_2019_arXiv,Verification_OOD_Detection_ODIN_Srikant_UIUC_ICLR_2018,Verification_SimpleFramework_OOD_Generative_Gaussian_NIPS_2018_7947}
study OOD and adversarial detection by setting a threshold
to decide if an input data is sufficiently similar to the pre-learnt  in-distribution
or non-adversarial data distribution.
These approaches do not offer a
measure of the actual
accuracy. Moreover, they
require access to the original training and/or validation
datasets, which are not needed by our accuracy
monitor. 

\textbf{Accuracy estimation for the target model.}
Secondary models are trained to estimate the accuracy of the primary model, but
they are trained on the same dataset as the primary model and requires 
either the
original input data \cite{mpp_2019} or saliency maps \cite{DNN_AccuracyPrediction_SaliencyMaps_Failures_AutonomousDriving_ZhangyangWang_TAMU_ICML_Workshop_2019_mohseni2019predicting}.
 In \cite{active_testing_accuacy_2018},  an active testing framework is
 proposed to estimate model accuracy,
  with a focus on
  noisy labeled datasets instead of
  unlabeled datasets
 that we consider.
Our problem is also related to operational testing \cite{DNN_Operational_Testing_NJU_FSE_2019_10.1145/3338906.3338930},
which uses selective random sampling
to provide an accuracy estimate for a target
DNN on an actual operational dataset prior to DNN deployment.
The work \cite{DNN_AccuracyPrediction_TAPAS_Architecture_Accuracy_Predictor_AAAI_2019_istrate2019tapas}
predicts the accuracy of a target DNN architecture on
a given dataset, while \cite{DNN_AccuracyPrediction_Weights_GoogleBrain_2020_unterthiner2020predicting}
predicts accuracy based on the target DNN's weights.
These studies
require a large number of DNN training experiments.

\textbf{Prediction confidence via softmax probability.}
A related study
\cite{DNN_Uncertainty_Baseline_OOD_ICLR_2017}
utilizes the maximum softmax probability of the target DNN for misclassification detection, whereas our approach exploits  the softmax probabilities for all classes.
Further, an abnormality module is designed to detect OOD data in \cite{DNN_Uncertainty_Baseline_OOD_ICLR_2017},
for which a decoder is required and trained with a white-box target model.
In \cite{DNN_Calibration_TemperatureScaling_ICML_2017_calibration_guo_2017}, {temperature scaling} is proposed to calibrate the original softmax probability,
but a labeled validation set is required to learn the hyperparameter $T$.
Likewise, \cite{DNN_Uncertainty_PostHoc_Dirichlet_NIPS_2019_kull2019beyond}
advances the temperature scaling method by training
a sophisticated Dirichlet distribution for better confidence calibration.
These methods are sensitive to and do not transfer
well to a user's datasets with OOD/adversarial samples.
\section{Problem Formulation}
We consider a deployed target DNN model that performs classification tasks
with $C$ classes. The DNN provides softmax probabilities
denoted as ${\mathbf{p}(x)}=M_{\Theta_d}(x)$, where $x$ represents the input data, ${\Theta}_d$ denotes target DNN's parameters  (not required by the accuracy monitor),  and $\mathbf{p}(x)\in \mathbb{R}^C$. Thus, the
predicted class is  $\tilde{y}={\arg \max}_{k\in\{1,2...C\}}\{p_k(x)\}$. 
The empirical accuracy $Acc$ of a deployed DNN model $M_{\Theta_d}$ on a user's dataset $(x_i,y_i)\in\mathcal{D}^U$ can be calculated as follows %Eqn.~\eqref{eqn:true_accuracy}.
\begin{equation}\label{eqn:true_accuracy}
Acc = \frac{1}{|\mathcal{D}^U|}\sum_{(x_i,y_i)\in\mathcal{D}^U}\mathbf{I}\left(y_i=\tilde{y}_i\right),
\end{equation}
where $\mathbf{I}(\cdot)$ is the Boolean indicator function.
The exact value of $Acc$ cannot be possibly obtained without
knowing all the true class $y_i$, which is often the case in practice
(e.g., a user employs a classifier due to the high cost of manually
labeling its data). It can also significantly
differ from the accuracy value evaluated based on
the DNN model provider's test dataset due to data
distribution disparity.

In this paper,
we leverage a simple plug-in accuracy monitor model to estimate the  empirical accuracy $Acc$ without all
the true labels for user's dataset. Specifically,
the neural network-based monitor model  $s(\mathbf{p}(x))=M_{\Theta_a}(\mathbf{p}(x))$ parameterized by $\Theta_a$ takes the target
DNN's softmax probabilities $\mathbf{p}(x)=M_{\Theta_d}(x)$
 as its input and outputs
a softmax probability/score $s(\mathbf{p}(x))$ to indicate the likelihood of correct classification for data $x$.
Then, if the probability of correct classification
is greater than or equal to a threshold $th_s$,
the target DNN's classification is considered correct and otherwise wrong.
By default,
we use $s(\mathbf{p}(x))\geq th_s=0.5$ in order
for a classification result to be considered correct.
Thus, the accuracy of the deployed DNN on the user's dataset estimated by our monitor model
is
\begin{equation}\label{eqn:estimate_accuracy}
\widetilde{Acc} = \frac{1}{|\mathcal{D}^U|}\sum_{(x_i,y_i)\in\mathcal{D}^U}
\mathbf{I}\left[s(\mathbf{p}(x))\geq th_s\right].
\end{equation}

Our problem formulation is similar
to that for the existing
confidence calibration techniques \cite{DNN_Uncertainty_PostHoc_Dirichlet_NIPS_2019_kull2019beyond,DNN_Calibration_TemperatureScaling_ICML_2017_calibration_guo_2017}
that focus on estimating the probability
of correct/wrong prediction for each \emph{individual}
sample.
Nonetheless, our key goal is to make the estimated \emph{average} accuracy
$\widetilde{Acc}$ as close to the true  empirical accuracy $Acc$ as possible.
%estimate the accuracy as accurately
This allows the application
of our method in even OOD/adversarial datasets, while still offering
an important view of the \emph{average} accuracy performance of
the target DNN.

Note finally that  our accuracy monitoring method does not require
a white-box target DNN model and
can be  applied on top of the target DNN
to monitor its accuracy performance,
either by the DNN model provider
or by an end user (provided that it has access
to a relevant labeled dataset, not necessarily the target DNN's
training/validation dataset). 

\section{Design of DNN Accuracy Monitoring}\label{sec:methodology}

Fig.~\ref{fig:monitor_system_overview} illustrates the flow of
our DNN accuracy monitor, including three phases. First, monitor models
are pre-trained over a labeled dataset that shares the
same application as the user's dataset.
Then, monitor models are re-trained with a small $t\%$ of labeled data from the user's dataset using active learning. Finally,
multiple monitor models are provided to approximate Bayesian neural networks via MC dropout, achieving a more robust accuracy estimation. Algorithm~1 describes the steps of our proposed method. Next, we provide details of the three phases for accuracy monitoring.

\begin{algorithm}[!t]
	\caption{DNN Accuracy Monitoring}
	\textbf{Input: }
	A labeled dataset $\mathcal{D}^R$, user dataset $\mathcal{D}^U$, \cloudModel $M_{\Theta_d}(x)$, the MC dropout model number $B$,
data labeling budget $t\%$.\\
	1. Obtain softmax probabilities for $\mathcal{D}^R$ and $\mathcal{D}^U$.\\
	\text{\quad\quad}$\mathbf{p}^R(x) \gets M_{\Theta_d}(x)\text{ for }x \in \mathcal{D}^R$;\\
	\text{\quad\quad}$CW^R(x) \gets \mathbf{I}\left(\tilde{y}=y \right)\text{ for }(x,y) \in \mathcal{D}^R$;\\
	\text{\quad\quad}$\mathbf{p}^U(x) \gets M_{\Theta_d}(x)\text{ for }x \in \mathcal{D}^U$;\\
	2. Train \monitors with $\mathbf{{p}}^R$ and $CW^R$.\\
	\For{b = 1 to B}{
		Initialize  $\Theta_a^{(b)}$  for a monitor model $M_{\Theta_a^{(b)}}$;\\
		Train $M_{\Theta_a^{(b)}}$ with $({\mathbf{p}}^R(x), CW^R(x))$;\\
		 $s^{(b)}(\mathbf{p}^U(x))\gets M_{\Theta_a^{(b)}}(\mathbf{p}(x))$ for $x\in\mathcal{D}^U$;
	}
	3. Actively label dataset $\mathcal{D}_s^U$ from user's dataset $\mathcal{D}^U$\\
	\text{\quad\quad}Calculate Shannon entropy $E(x)$ based on\\
\text{\quad\quad}$s^{(b)}(\mathbf{p}^U(x))$ and average over $B$ monitor\\
\text{\quad\quad} models  for
$(x,y)\in \mathcal{D}^U$;\\
    \text{\quad\quad}$\mathcal{D}_s^U \gets \left\{ (x,y)\in\mathcal{D}^U | E(x) \text{ among the top } t\% \right\}$;\\
	\text{\quad\quad}$\mathbf{{p}}_s^U(x) \gets M_{\Theta_d}(x)\text{ for }(x,y) \in \mathcal{D}_s^U$;\\
	\text{\quad\quad}$CW_s^U(x) \gets \mathbf{I}\left(\tilde{y}=y \right)\text{ for }(x,y)\in \mathcal{D}_s^U$;\\
	4. Transfer learning and accuracy estimation. \\
	\For{b = 1 to B}{
		Transfer $M_{\Theta_a^{(b)}}$ with  $(\mathbf{p}_s^U(x), CW_s^U(x))$;\\
		$s^{(b)}(\mathbf{p}^U(x))\gets M_{\Theta_a^{(b)}}(\mathbf{p}^U(x))$ for $x\in\mathcal{D}^U
\setminus \mathcal{D}_s^U$;
}
	\Return Average $\widetilde{Acc}$ from Eqn.~\eqref{eqn:estimate_accuracy};
	\label{algo:monitor_algorithm}
\end{algorithm}

\textbf{Training phase.}
To pre-train initial monitor models,
the accuracy monitor
can leverage a labeled dataset $\mathcal{D}^R$,
which can be the target DNN's training/validation dataset
(if the DNN provider wants to monitor its own model's
accuracy)
or a different dataset relevant to the target application
(if the DNN user wants to monitor the
accuracy by itself but does not have the target DNN's original
training/validation dataset).
For example, if the target
DNN is developed by one entity
but later provided to another user as a black-box model
for image classification,
 CIFAR10, CINIC10 or ImageNet2012 can be used by
 the user to pre-train its own accuracy monitor models. We run
  the target DNN on the labeled dataset and obtain prediction softmax probabilities $\mathbf{p}^R(x)$
 produced by the target DNN. Meantime, the correct/wrong
 result $CW^R(x)$  of the target DNN
 can also be obtained by comparing the DNN's predicted class
 with the true data label. Then, based on $\mathbf{p}^R(x)$ and  $CW^R$,
 we can train $B$ monitor models $M_{\Theta_a^{(b)}}$.

\textbf{Transfer with active learning. } Due to the possible distribution differences between the chosen labeled dataset $\mathcal{D}^R$ and the user's actual dataset $\mathcal{D}^U$, the
monitor models pre-trained solely on the $\mathcal{D}^R$ may not provide a satisfactory accuracy for the target
DNN as shown in Section~\ref{sec:experiment}.
To address this issue,
we need to transfer the monitor models into  the user's dataset.
In the transfer learning phase, we freeze the weights of all layers in the monitor models except for the last two layers. Only the weights of the last two layers will be updated during transfer learning. Due to expensive labeling cost, we only sample a small amount
of user's dataset (denoted as $\mathcal{D}_s^U$) from
 $\mathcal{D}^U$, and only $\mathcal{D}_s^U$ are manually labeled. To minimize the size of $\mathcal{D}_s^U$, entropy-based active learning \cite{ensemble_active_learning_2018} is utilized during the transfer. {Specifically, we calculate the average entropy of softmax probabilities produced by the monitor models, and
 label $t\%$ of user's data with the greatest entropy.}

 Note that while labeling user's data, only the user's data label $y$ and deployed DNN's softmax probabilities $\mathbf{p}(x)$
 (instead of the raw data $x$) are utilized
 by the monitor models.
 Moreover, by doing so, the accuracy monitor actually performs accuracy estimation
 of the target DNN model
 over a low-dimension softmax probability representation of $x$,
 which effectively facilities transfer learning to user's dataset. As shown in
 our experiments,
 by labeling only 1\% of the user's dataset, the monitor models can produce a highly accurate estimation of the target DNN's average accuracy.

\textbf{Robust accuracy estimation with MC dropout.} Estimating accuracy for the target DNN by a single monitor model may not be robust because of the indispensable uncertainty in deep learning.  Based on \cite{MC_dropout_approximation_2016}, we employ
the  MC dropout method to approximate a Bayesian neural network and provide more robust accuracy estimation. 
Specifically, we train an ensemble of monitor models in the training phase using the same labeled dataset but different initialized weights and dropout layers. Then, we transfer the trained models using the same dataset $\mathcal{D}_s^U$. When estimating the target DNN's classification accuracy, multiple estimated accuracies can be obtained from the ensemble. The mean of the results is considered as the monitor's
assessment on the deployed DNN's classification accuracy over the user's dataset.
Moreover,
the standard deviation (std) can also be provided to represent the uncertainty of estimated accuracy by the ensemble of monitor models.
\begin{table*}[!t]
	\centering
	\caption{Performance of our method and baseline algorithms on 10-class image classification. The mean/std values are provided for our method. Target DNN: VGG16 trained on CIFAR-10}
	\setlength{\tabcolsep}{5pt}
	\vspace{-0.1cm}
	\small

	\begin{tabular}{l|cccc|cccc}
		\hline
		\multirow{2}{*}{\textbf{Method}} & \multicolumn{4}{c|}{\textbf{Estimated Accuracy}} & \multicolumn{4}{c}{\textbf{AUPR}} \\
		\cline{2-9}      & CIFAR-10 & CINIC-10 & STL-10 & AD-10 & CIFAR-10 & CINIC-10 & STL-10 & AD-10 \\
		\hline
		\multicolumn{1}{l|}{\textbf{Our method}} & \textbf{0.9313}/0.0123 & \textbf{0.7691}/0.0138 & \textbf{0.6343}/0.0371 & \textbf{0.3866}/0.0322 & 0.9270 & \textbf{0.8645} & \textbf{0.7966} & \textbf{0.8935} \\
		\multicolumn{1}{p{5.5em}|}{\textbf{MP}} & 0.8907 & 0.7574 & 0.7105 & 0.5035 & 0.9341 & 0.8595 & 0.7922 & 0.8918 \\
		\multicolumn{1}{p{5.5em}|}{\textbf{Entropy}} & 0.8943 & 0.7662 & 0.7165 & 0.5380 & \textbf{0.9352} & 0.8645 & 0.7966 & 0.8859 \\
		\textbf{TS} & 0.9727 & 0.4066 & 0.8803 & 0.8618 & 0.9343 & 0.8607 & 0.7964 & 0.8922 \\
		\multicolumn{1}{p{5.5em}|}{\textbf{MP*}} & 0.9756 & 0.9443 & 0.9319 & 0.7881 & -     & -     & -     & - \\
		\textbf{RS (1\%)} & [0.8879,0.9852] & [0.6500,0.7340] & [0.5274,0.7382] & [0.2800,0.5100] & -     & -     & -     & - \\
		\textbf{RS (10\%)} & [0.9207,0.9516] & [0.7340,0.7930] & [0.5976,0.6618] & [0.3400,0.4080] & -     & -     & -     & - \\
		\hline
	\end{tabular}%
\label{tab:DNN_10classy}%
\vspace{-0.3cm}
\end{table*}%

\section{Experiments}\label{sec:experiment}
We first evaluate
the effectiveness of our accuracy monitoring method
on two image classification applications:  small-scale  image classification with 10 classes, and large-scale  image classification with 1000 classes. Then, we consider a mission-critical application --- traffic sign detection for autonomous driving.

\subsection{Setup}

Our accuracy monitor model is trained as a neural network with dropout layers using Tensorflow  and Keras \cite{tensorflow_2016}. %\cite{keras_2015}.
The weight parameter $\Theta_a$ is trained via minimizing binary cross-entropy loss using Adam \cite{adam_sgd_2014} with a learning rate $\alpha=0.001$.
The input of the monitor model  is the softmax probabilities $\mathbf{p}(x)$ produced by the target DNN, while the output represents if
the classification is correct or not for an input image $x$ with a softmax score $s(\mathbf{p(x)})$, which will then be averaged over multiple
samples to form an estimate of the average accuracy.

\textbf{Dataset.} The datasets include CIFAR-10 \cite{cifar10_dataset}, CINIC-10 \cite{cinic10_dataset}, STL-10 \cite{stl10_dataset}, ImageNet2012 \cite{imagenet_2012} \zhihui{and German Traffic Sign Detection (GTSD) \cite{traffic_sign_dataset_IJCNN_2013}}. In addition, we also consider a user's dataset with adversarial images for 10-class classification \zhihui{and GTSD  classification, denoted as AD-10, and GTSD-AD, respectively}. The adversarial images are generated using DeepFool  \cite{deepfool_adversarial_attack_2016} policy with {``Foolbox''} package \cite{fool_box_2017}.

\textbf{Target DNN model.}
The target DNN model for 10-class image classification is VGG16 \cite{vgg_16_2014}, while MobileNet \cite{mobilenets_2017_Howard} and ResNet-50 \cite{resnet50_2016} are used as the target DNNs for 1000-class image classification. \zhihui{The target model for GTSD is a native convolutional neural network (CNN) trained on the GTSD training dataset.} The accuracy monitor estimates the
classification accuracy achieved by these DNNs
on the above datasets
 (which can be OOD with respect
to the DNNs' original training datasets).

\subsection{Baseline Approaches and Metrics}
The following baselines and metrics are considered.

\textbf{RS:} With random sampling (RS), $u\%$ of user's data is  randomly sampled and manually labeled. Then, the accuracy on the sampled user's dataset is considered as the overall accuracy.
We also run RS for 100 times, and highlight the  accuracy range achieved by 100 runs. Note, however, that in practice the RS is only performed {once} for each test dataset.

\textbf{MP and MP*:} In the MP approach considered
in \cite{DNN_Uncertainty_Baseline_OOD_ICLR_2017}, no manual
labeling is needed; instead,
the maximum softmax probability
$MP(x)={\max}_{k\in\{1,2...C\}}\{p_k(x)\}$
produced by the target DNN model
is utilized: if $MP(x)\geq th_{MP}$
where $th_{MP}$ is a threshold, then the classification for $x$ is considered correct and otherwise wrong.
In our experiment, the threshold $th_{MP}$ is determined
based on the same labeled dataset used to train our monitor models
to achieve the best accuracy estimation during the validation.
Alternatively, we can also use the maximum softmax probability
on the use's dataset to estimate the target DNN's accuracy as
$MP$*$=\sum_{(x,y)\in\mathcal{D}^U}MP(x)$, and we use MP* to represent
this approach.

\textbf{Entropy:} The prediction entropy $Entropy(\mathbf{p}(x))$ can be calculated from softmax probability of the target DNN model. Then, the target DNN's classification for $x$ is considered correct if $Entropy<th_{En}$, where $th_{En}$ is the entropy threshold decided by the monitor according
to its chosen labeled ataset, and wrong otherwise.

\textbf{Temperature scaling (TS)}: By using temperature scaling, the softmax probability can be calibrated from the logits  with a hyper-parameter $T$. According to \cite{DNN_Calibration_TemperatureScaling_ICML_2017_calibration_guo_2017}, given the logit output $z_i$, the model accuacy can be estimated as $TS=\max{\sigma_{SM}(z_i/T)}$, where $\sigma_{SM}$ is the softmax function and $T$ is called the temperature. Usually, the temperature $T$ is obtained via minimizing the Negative log likelihood (NLL) on the target
DNN's validation set. Here, we use the actively labeled
user's data samples as the validation set.

\textbf{Perfect confidence calibration}:
This is an oracle that gives the true accuracy of the target
DNN and no practical confidence calibration
methods (e.g., \cite{DNN_Uncertainty_PostHoc_Dirichlet_NIPS_2019_kull2019beyond}) can outperform.

\textbf{Metrics:} Our main performance metric is the estimated average accuracy of the target DNN. Additionally, we also consider AUPR (Area Under the  Precision-Recall Curve) to isolate the effects of different thresholds $th_s$.
The value of threshold-less AUPR varies from positive class ratio $p$ (random guess) to 1.0 (perfect classification), and measures a model's capability of distinguishing between correct/wrong classification. The higher AUPR, the better.

\subsection{Result on 10-class Image Classification}
\begin{figure}[!t]
	\centering
	\subfigure[]{\includegraphics[trim=0cm 0cm 0cm 0cm,clip,  width=0.175\textwidth]{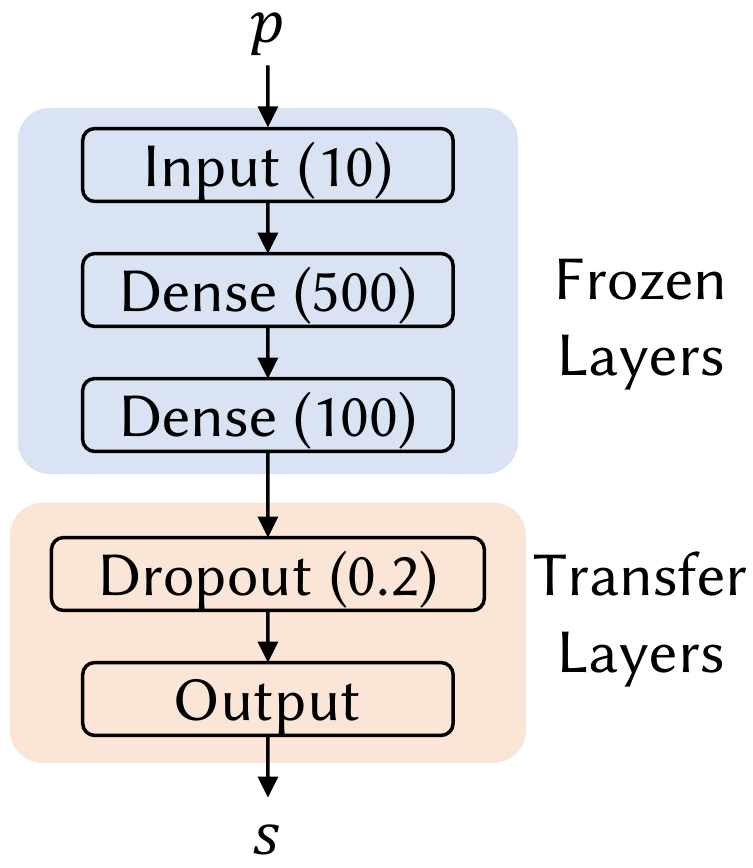}\label{fig:cifar10_model}}
	\subfigure[]{\includegraphics[trim=0cm -1cm 0cm 0cm,clip,  width=0.30\textwidth]{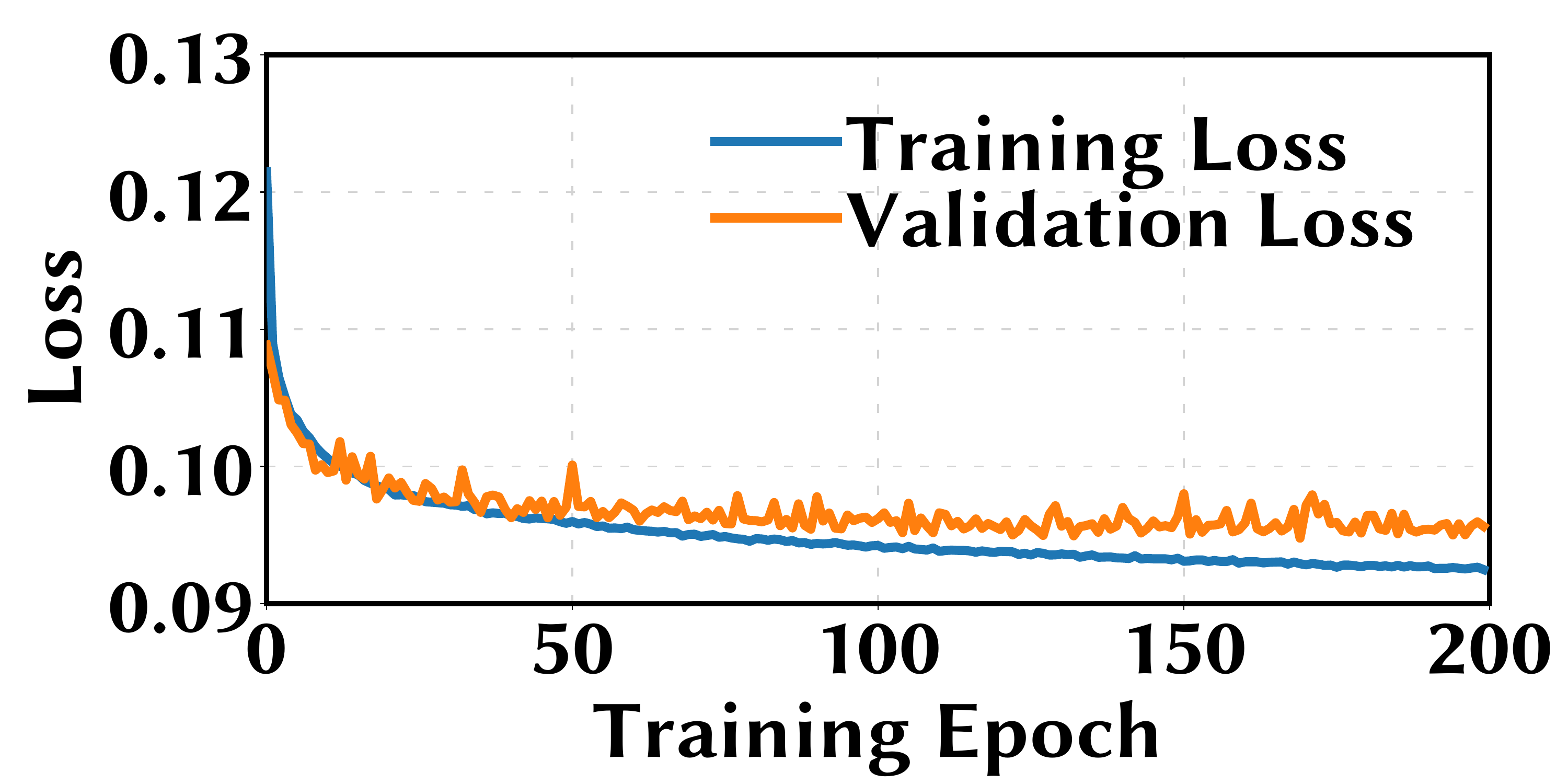}\label{fig:cifar10_model_loss}}
	\vspace{-0.3cm}
	\caption{(a) The \monitorModel structure for 10-class image classification. (b) Loss during \monitorModel training. }\label{fig:cifar10_model_train}
	\vspace{-0.3cm}
\end{figure}

For 10-class image classification, the \cloudModel is a VGG16 model trained on CIFAR-10 \cite{vgg16_weights}. \zhihui{We evaluate the performance of our proposed method on four datasets shown in Table~\ref{tab:DNN_10classy}. The dataset sizes are 10k (CIFAR-10), 90k (CINIC-10), 8k (STL-10) and 10k (AD-10). The reported inference accuracy of the target VGG16 model is 93.56\% measured on CIFAR-10, while the inference accuracies for other datasets are 76.17\% (CINIC-10), 63.04\% (STL-10), and 37.80\% (AD-10), indicating a significant accuracy degradation due to OOD/adversarial data.} First, we train an ensemble of 20 \monitors on 9000 images from a public dataset
(i.e., CINIC-10 training dataset in our experiment) and the structure of \monitorModel is shown in Fig.~\ref{fig:cifar10_model}, including two hidden dense layers and one dropout layer.

In the training phase, each \monitorModel is trained over 200 epochs with Adam optimizer. Fig.~\ref{fig:cifar10_model_loss} shows the training and validation loss for a monitor model in training phase. Then, two hidden layers are frozen to perform transfer learning as shown in Fig.~\ref{fig:cifar10_model}. To improve the transfer efficiency, an active learning approach is utilized to select 1\% samples with the highest
entropy from the user's test dataset.
In the prediction phase,  robust estimation and its uncertainty are provided by the ensemble of monitor models.

\begin{table*}[!t]
	\centering
	\caption{Performance of our method and baseline algorithms on 1000-class image classification. The mean/std values are provided for our method. Target DNN: MobileNet/ResNet-50 model.}
	\setlength{\tabcolsep}{2.5pt}
	\vspace{-0.1cm}
	\small
	\begin{tabular}{l|cccc|cccc}
		\hline
		\multirow{3}{*}{\textbf{Method}} & \multicolumn{4}{c|}{\textbf{Estimated Accuracy}} & \multicolumn{4}{c}{\textbf{AUPR}}\\
		\cline{2-9}      & \multicolumn{2}{c}{\textbf{MobileNet}} & \multicolumn{2}{c|}{\textbf{ResNet-50}} & \multicolumn{2}{c}{\textbf{MobileNet}} & \multicolumn{2}{c}{\textbf{ResNet-50}} \\
		\cline{2-9}      & ImageNet A & ImageNet B & ImageNet A & ImageNet B & ImageNet A & ImageNet B & ImageNet A & ImageNet B \\
		\hline
		\multicolumn{1}{p{5.5em}|}{\textbf{Our method}} & \textbf{0.6933}/0.0202 & \textbf{0.6796}/0.0235 & \textbf{0.6862}/0.0240 & \textbf{0.6719}/0.0219 & \textbf{0.7192} & \textbf{0.7245} & \textbf{0.7066} & \textbf{0.7175}\\
		\multicolumn{1}{p{5.5em}|}{\textbf{MP}} & 0.7203 & 0.7004 & 0.6765 & 0.6757 & 0.7182 & 0.7221 & 0.7050 & 0.7103 \\
		\multicolumn{1}{p{5.5em}|}{\textbf{Entropy}} & 0.7032 & 0.7131 & 0.6724 & 0.6694 & 0.7052 & 0.7015 & 0.6907 & 0.7050 \\
		\textbf{TS} & 0.8094 & 0.8086 & 0.7771 & 0.8044 & 0.7123 & 0.7197 & 0.7059 & 0.7150 \\
		\multicolumn{1}{p{5.5em}|}{\textbf{MP*}} & 0.7550 & 0.7539 & 0.7633 & 0.7638 & -     & -     & -     & - \\
		\textbf{RS (1\%)} & [0.6197,0.7512] & [0.5866,0.7754] & [0.6631,0.7029] & [0.6457,0.7049] & -     & -     & -     & - \\
		\textbf{RS (10\%)} & [0.6652,0.7057] & [0.6492,0.7073] & [0.6696,0.6987] & [0.6525,0.6959] & -     & -     & -     & - \\
		\hline
	\end{tabular}

	\label{tab:DNN_1000class}
	\vspace{-0.3cm}
\end{table*}

The estimated accuracy results are summarized in Table~\ref{tab:DNN_10classy}, compared with baseline approaches.
Our method can provide much more accurate
estimate of the target DNN's inference accuracy on user's test
datasets. While our monitor models are trained
on CINIC-10, with transfer learning on only 1\% of the user's dataset, we can still accurately estimate
the target DNN's inference accuracy when user's
dataset is STL-10.

The inference accuracy via the RS approach exhibits a large variance with 1\% labeled data, and at least $10\%$ labeled samples are required to achieve a small estimation error. For temperature scaling method, the estimated accuracy still deviates from the true accuracy with a large gap. For MP-based and entropy-based approaches, the estimated accuracy varies greatly with threshold values.
Although in theory one can always find a threshold with which
the resulting estimated accuracy coincides with the true accuracy $z\%=Acc$,
such a threshold is not very meaningful, since it simply says
samples with top  $z\%$ maximum softmax probability or entropy
are correct.

As shown in Table~\ref{tab:DNN_10classy}, our method has a similar AUPR value with the baseline
approaches, demonstrating that the
overall capability of distinguishing correct/wrong classification is
comparable among different methods. Nonetheless, AUPR is not
as
an intuitive metric as average accuracy, which our accuracy monitor is specifically designed for. Also, AUPR is only applicable for methods with variable thresholds (e.g., MP, Entropy and TS) as provided in Table~\ref{tab:DNN_10classy}.

\begin{figure}[!t]
	\centering
	\subfigure[]{\includegraphics[trim=0cm 1cm 0cm 0cm,clip,  width=0.23\textwidth]{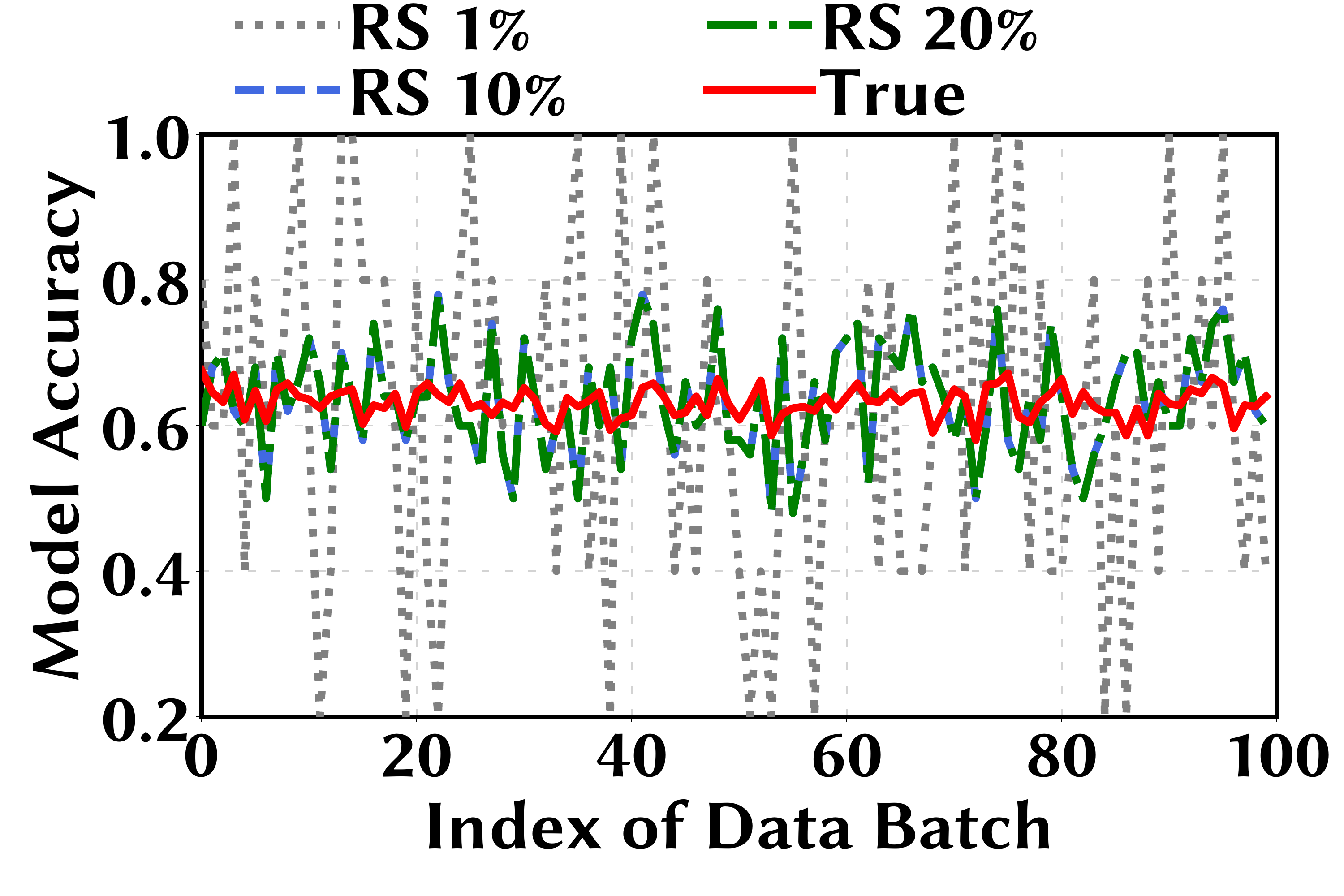}\label{fig:stl_stream_RS}}
	\subfigure[]{\includegraphics[trim=0cm 1cm 0cm 0cm,clip,  width=0.23\textwidth]{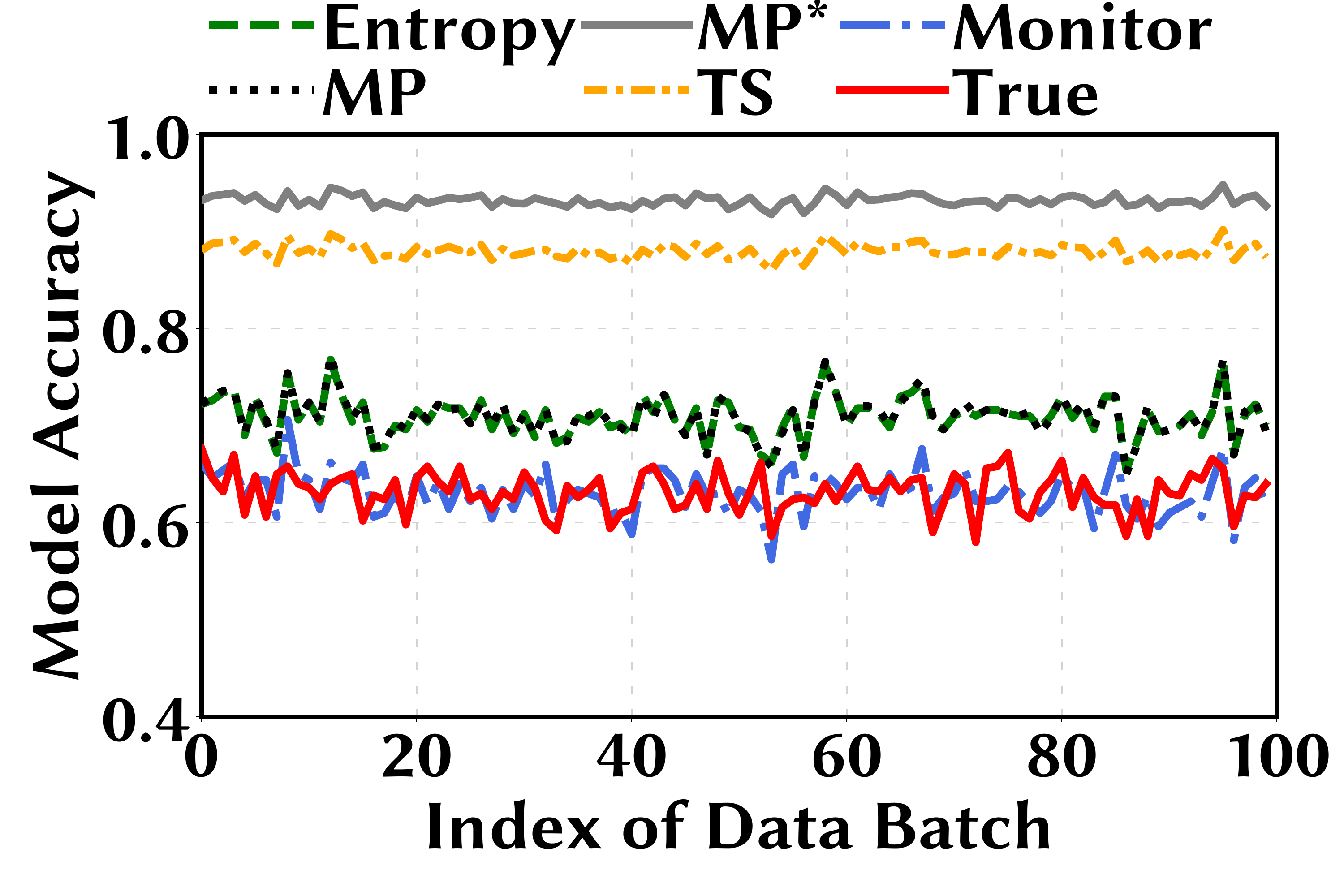}\label{fig:stl_stream_auditor}}
	
	\vspace{-0.3cm}
	\caption{Performance evaluation for different algorithms on batched datasets from STL10, each including 500 images. }\label{fig:stream_data}
	\vspace{-0.3cm}
\end{figure}

In addition, we also evaluate the performance of \monitorModel on
small-batch datasets to see if our monitor models
can track the true empirical accuracy of the target DNN
on user's time-varying datasets. Specifically, we randomly select
500 images as a batch from STL-10 with replacement, and
repeat to have a total of 100 batches each having 500 images.
We show the results in Fig.~\ref{fig:stream_data}
and demonstrate our accuracy monitor can closely track
the empirical true accuracy, whereas the baseline approaches
cannot. Even 20\% RS (i.e., randomly label 100 images
for each batch) {and temperature scaling algorithms} cannot provide a good accuracy estimate.

\subsection{Result on 1000-class Image Classification}
For image classification with 1000 classes, two \cloudModels (MobileNet and ResNet-50) are applied on ImageNet2012's validation dataset. The original validation set includes 50k images. We randomly split ImageNet2012 into 3 datasets: training dataset with 20k, ImageNet A with 20k images and ImageNet B with 10k images. The reported accuracies on ImageNet dataset are 70.40\% for MobileNet and 74.90\% for ResNet-50, respectively. For MobileNet, the true accuracies on test datasets are 68.59\% (ImageNet A) and 67.91\% (ImageNet B), respectively. For ResNet-50, the true accuracies are 68.36\% (ImageNet A) and 67.47\% (ImageNet B), respectively. The true accuracies vary due to the  distribution shift.

For the 1000-class \cloudModel, the softmax probability $\mathbf{p}(x)$ includes 1000 values. Therefore, the \monitorModel structure is changed accordingly with 1000 input nodes and 1000 hidden nodes in hidden layers. Other settings remain the same. The results for MobileNet and ResNet-50 on ImageNet-A and ImageNet-B are summarized in Tables~\ref{tab:DNN_1000class}. They demonstrate that the \monitorModel also outperforms baseline approaches for large-scale image classification.
Similarly, the RS's estimated accuracy exhibits a high variation and at least 10\% labeled data are required to achieve a similar performance as the \monitorModel.
Due to distribution similarity
between the training dataset and ImageNet A/B which
are all selected from ImageNet2012, the MP-based
and {entropy-based} approaches (with thresholds optimized
based on the training dataset) offer a reasonable estimate of
the true accuracy, but they are still worse than our monitor model.
Similarly, temperature scaling has a higher estimation error due to limited (1\%) labeled samples.
Our accuracy monitor exhibits a
slightly
larger estimation error on 1000-class models than the 10-class case. One possible reason is the higher dimensions in the softmax probability, which may require more complex feature extraction layers instead of  simple fully-connected layers
in our current experiment.

\begin{table*}[!t]
	\centering
	\caption{Performance of our method and baseline algorithms on traffic sign detection. The mean/std values are provided for our method. Target DNN: a CNN model trained on GTSD.}
	\setlength{\tabcolsep}{3.5pt}
	\vspace{-0.2cm}
	\small
	\begin{tabular}{l|cccc|cccc}
		\hline
		\multirow{2}{*}{\textbf{Method}} & \multicolumn{4}{c|}{\textbf{Estimated Accuracy}} & \multicolumn{4}{c}{\textbf{AUPR}} \\
		\cline{2-9}      & GTSD-D1 & GTSD-D2 & GTSD-OOD & GTSD-AD & GTSD-D1 & GTSD-D2 & GTSD-OOD & GTSD-AD \\
		\hline
		\multicolumn{1}{p{5.5em}|}{\textbf{Our method}} & \textbf{0.9735}/0.001 & \textbf{0.8414}/0.005 & \textbf{0.5362}/0.005 & \textbf{0.4162}/0.001 & \textbf{0.6585} & \textbf{0.6955} & \textbf{0.9090} & 0.8414 \\
		\multicolumn{1}{p{5.5em}|}{\textbf{MP}} & 0.9837 & 0.7991 & 0.5690 & 0.4886 & 0.6106 & 0.6026 & 0.8873 & 0.8414 \\
		\multicolumn{1}{p{5.5em}|}{\textbf{Entropy}} & 0.9621 & 0.7866 & 0.5821 & 0.4806 & 0.6248 & 0.6027 & 0.8952 & \textbf{0.8471} \\
		\textbf{TS} & 0.9855 & 0.8004 & 0.7157 & 0.6884 & 0.6211 & 0.5765 & 0.8994 & 0.8094 \\
		\multicolumn{1}{p{5.5em}|}{\textbf{MP*}} & 0.9895 & 0.9390 & 0.8861 & 0.9176 & -     & -     & -     & - \\
		\textbf{RS (1\%)} & [0.9406,1.000] & [0.7624,0.9307] & [0.4300,0.5900] & [0.3533,0.4900] & -     & -     & -     & - \\
		\textbf{RS (10\%)} & [0.9574,0.9871] & [0.8178,0.8693] & [0.4880,0.5387] & [0.4100,0.4460] & -     & -     & -     & - \\
		\hline
	\end{tabular}
	\label{tab:GTSDclass_accuracy}
	\vspace{-0.3cm}
\end{table*}

\subsection{Result on Traffic Sign Detection}
We now consider traffic sign detection in safety-critical autonomous driving on the GTSD dataset, including 40k samples grouped into 43 categories/classes \cite{traffic_sign_dataset_IJCNN_2013}. We train a CNN on GTSD training dataset (27k samples) using 50 epochs via Adam optimizer. The CNN includes convolution layers, dropout layers, and fully connected layers.

We evaluate the proposed method and baseline approaches on four test datasets generated from GTSD, including the original test dataset (GTSD-D1), augmented test dataset (GTSD-D2), out-of-distribution dataset (GTSD-OOD), and adversarial dataset (GTSD-AD). Specifically, GTSD-D1 includes 10k samples randomly selected from the GTSD test dataset, while GTSD-D2 includes 10k augmented samples from the GTSD test dataset. The augmentation operations and parameters for GTSD-D2 are random rotation within $[-10,10]$ degrees and random vertical/horizontal shift within $[-0.1,0.1]$. As for GTSD-OOD, it includes 12k OOD samples from CIFAR-10 and 18k samples from the augmented dataset GTSD-D2. The OOD samples from CIFAR-10 are resized into $(30,30,3)$ using tf.image.resize function with default parameters, and they are treated with NULL label, indicating not belonging to any of the 43 classes in GTSD. The GTSD-AD dataset includes 15k normal samples and 15k adversarial samples. The normal samples are randomly selected from the augmented dataset GTSD-D2, while adversarial samples are generated with DeepFool. The reported inference accuracy measured on GTSD-D1 is 97.34\%, while the inference accuracies for the other datasets are 84.01\% (GTSD-D2), 51.47\% (GTSD-OOD) and 42.91\% (GTSD-AD), respectively.

For the target DNN, the softmax probability vector $\mathbf{p}(x)$ includes 43 elements. The structure of our monitor model in Fig.~\ref{fig:cifar10_model} is modified to
include 100 and 50 hidden nodes in two hidden layers, respectively. First, we pre-train an ensemble of 20 monitor models on a public dataset (for which we choose GTSD-D2 in our evaluation). In the training phase, each \monitorModel is trained over 200 epochs with Adam optimizer. Then, when applied to different datasets, the weights in the first two layers are frozen to perform transfer learning. Other settings remain the same.

The estimated accuracies by different methods  are summarized in Table~\ref{tab:GTSDclass_accuracy}. The results show that our method still outperforms
the considered baselines, providing a much more accurate estimate of the target DNN's inference accuracy on user's datasets. With pre-trained monitor models and only 1\% labeled data, we can accurately estimate the target DNN's inference accuracy when applied to different datsets (GTSD-OOD or GTSD-AD). Also, the estimated accuracy by RS exhibits a high variation with 1\% labeled data and at least 10\% labeled data is required to achieve a similar performance as our method. Additionally, the baselines provide an estimated accuracy with a large error. For instance, the MP* and TS methods often provide a higher estimated accuracy than the true accuracy.

To sum up, the results on GTSD further demonstrates the effectiveness of our proposed method for accuracy estimation of a target DNN.

\section{Conclusion}
{In this paper, to increase
the trustworthiness of DNN classification results,
we propose a post-hoc method for monitoring the prediction performance of a target DNN models and estimating its empirical inference accuracy
on user's (possibly OOD/adversarial) dataset.}
The monitor model only takes the softmax probability produced by the target DNN model as its input. Thus, it can be easily employed as a plug-in module on top of a target DNN to monitor its accuracy.
Importantly, by active learning with a small amount of labeled data from user's datasets, our monitor model can produce a very accurate estimate of inference accuracy of the target DNN model. Our experiment results on different datasets validate the effectiveness and efficiency of the proposed method for image classification and traffic sign detection.

\section*{Acknowledgments}
This work was supported in part by the U.S. National Science Foundation under grants CNS-1551661, ECCS-1610471, and CNS-1910208.

\bibliographystyle{named}

\end{document}